\documentclass[11pt]{article}
\usepackage[letterpaper,margin=1in]{geometry}
\usepackage{graphicx}
\usepackage[T1]{fontenc}

\usepackage{amsmath}
\usepackage{amsfonts}
\usepackage{amssymb}
\usepackage{array}

\usepackage{listings}
\usepackage{xcolor}

\newcommand{\greencheck}{\textcolor{green!60!black}{\checkmark}}

\definecolor{codegray}{rgb}{0.5,0.5,0.5}
\definecolor{codepurple}{rgb}{0.58,0,0.82}
\definecolor{codegreen}{rgb}{0,0.5,0}
\definecolor{backcolour}{rgb}{0.97,0.97,0.95}

\lstdefinestyle{pythonstyle}{
    backgroundcolor=\color{backcolour},
    commentstyle=\color{codegreen},
    keywordstyle=\color{blue},
    numberstyle=\tiny\color{codegray},
    stringstyle=\color{codepurple},
    basicstyle=\ttfamily\footnotesize,
    breaklines=true,
    captionpos=b,
    keepspaces=true,
    numbers=left,
    numbersep=4pt,
    showspaces=false,
    showstringspaces=false,
    showtabs=false,
    tabsize=2,
    language=Python
}
\usepackage[square,numbers]{natbib}

\usepackage[colorlinks=true,urlcolor=blue,citecolor=black,
            anchorcolor=black,linkcolor=black]{hyperref}

\begin{document}

\title{SupplyNetPy: An Open-Source Python Library for High-Fidelity Modeling
and Simulation of Arbitrary Supply Chain and Inventory Networks}

\author{%
Tushar Lone\textsuperscript{1} and Neha Karanjkar\textsuperscript{1}\\[4pt]
\textsuperscript{1}School of Mathematics and Computer Science,\\
Indian Institute of Technology Goa, Ponda, Goa, INDIA \\[6pt]
\textit{\small This is the author's preprint of a paper accepted at Winter
Simulation Conference, 2026.} 
}

\date{}

\maketitle

\begin{abstract}
This paper introduces SupplyNetPy, an open-source, well-documented Python
library for modeling and discrete-event simulation of supply chain networks
with arbitrary multi-echelon structures. It supports multiple
replenishment policies, perishable inventory, node disruptions, and stochastic
demand and lead times. All components are extensible via inheritance. Users
describe a supply chain as a graph with node and link attributes, while the
library handles simulation, providing logs and extensive node and network level
performance reports. This paper presents the motivation, design, key features,
and architecture of SupplyNetPy, along with detailed validation results
(against analytical benchmarks, a commercial tool, and a published case study).
A key motivation behind SupplyNetPy's development is programmatic generation
and simulation of complex models, enabling design-space exploration, what-if
analysis, training data generation, and supply chain digital twins.
\end{abstract}


\section{Introduction}
\label{sec:intro}

Modern supply chains (SCs) are complex, dynamic networks that span the globe and are
acutely vulnerable to disruptions. Their design, optimization, and management
under continuous fluctuations requires analysis tools that can handle 
complexity and assimilate real-time information.
The recent 2026 geopolitical situation and disruptions to shipping lanes (such as in the
Strait of Hormuz) have sent far-reaching shockwaves through the global economy. Commodity prices, shipping lead times, and inventory costs shifted in ways that were hard to predict. This has acutely highlighted the need for robust SC analysis and management tools.
The intricate behavior, global scale, and stochastic nature of modern SCs
limit the scope of purely analytical models, spreadsheets, or human intuition
for their management. Simulation modeling has become indispensable and assimilation of real-time data into the model (that is, a simulation-based digital twin of SCs) has become both feasible and essential \citep{biller2023simulation,lugaresi2023digital}.

Commercial simulation tools such as
\href{https://www.arenasimulation.com/}{Arena} and
\href{https://www.anylogic.com/}{AnyLogic} are commonly used for modeling and
analyzing SCs, particularly in areas like inventory management,
logistics, and resilience. These tools offer generic components for modeling
discrete-event systems, such as processes and shared resources, but may not
have component libraries that are specific to SCs. Examples of tools
that are specific for SC modeling are
\href{https://www.anylogistix.com}{AnyLogistix} and
\href{https://github.com/amzn/supply-chain-simulation-environment/}{miniSCOT}, which provide SC-specific
components.
However, very few open-source libraries exist that are specifically targeted
for SCs. These include
\href{https://github.com/KevinFasusi/supplychainpy}{\textit{supplychainpy}},
miniSCOT, and \href{https://github.com/aitechtools/SunFlow}{SunFlow}.
%
The open-source alternatives do not support arbitrary networks (single-echelon only). Several do not support simulation 
(using analytical methods and spreadsheets) or are not well-maintained, making them difficult to consider as viable
alternatives to commercial tools. A detailed review of existing tools is presented in Section~\ref{sec:related}.
There is a dearth of well-maintained, well-documented open-source
libraries specifically targeted for the discrete-event simulation (DES) of arbitrary SC networks.
Practitioners often fall back on general-purpose DES frameworks such as
\href{https://simpy.readthedocs.io/}{SimPy} and build all SC logic
from scratch \citep{nam2022development}, which demands considerable development and validation effort. 
An open-source library providing ready-to-use, composable SC building blocks reduces this
burden, supports community-driven maintenance and extensibility, and enables
seamless integration with Python's rich ecosystem for optimization, machine
learning, and data analysis. This motivates our development of SupplyNetPy.

\subsection{An Overview of SupplyNetPy and its Features}
\label{subsec:overview}

SupplyNetPy is an open-source Python library for expressive modeling and
DES of SC networks. It is installable from
the Python Package Index via \texttt{\ pip install supplynetpy\ } with detailed documentation, 
user guides and full examples at \url{https://supplychainsimulation.github.io/SupplyNetPy},
and source code on GitHub \citep{supplynetpy_github}.
It has been built as a component library using
\href{https://simpy.readthedocs.io/}{SimPy} at its core for performing
event-driven simulation \citep{simpy}.

A SC is described by the user as a graph: nodes represent SC 
entities and links represent transportation connections between them.
\textbf{Node attributes} (such as inventory capacity, replenishment policy, reorder levels,
failure distributions for disruption modeling, and shelf life for perishable inventory)
and \textbf{link attributes} (such as transportation cost, and lead time distributions)
can be specified by the user. The library handles all simulation mechanics, and provides 
detailed event logs and node and network level performance statistics.
Key features include support for disruption modeling and resilience analysis.
Perishable inventory is supported with per-unit expiry tracking using a
first-in, first-out (FIFO) discipline and waste cost accounting, making the
library attractive for use cases such as food SCs, milk distribution
networks, cold chains for vaccines, and pharmaceutical SCs. The
library also provides several built-in inventory replenishment policies and
supplier selection strategies, all extensible via inheritance. Simulations
automatically generate extensive node and network level performance reports,
covering inventory levels, shortages, waste, costs, revenue, and service
metrics.

The library has been thoroughly validated through comparison with analytical
benchmarks, against a commercial simulation tool (AnyLogistix) for unit-tests with deterministic configurations, 
and against published results from a case study. The design, architecture,
and detailed validation results are presented in
Sections~\ref{sec:library} to \ref{sec:casestudy}.

Aside from the general benefits of open-source development (community
contributions, long-term maintainability, and transparency), the following
were key motivations behind the development of SupplyNetPy.

\begin{itemize}

    \item \textbf{Programmatic model generation at scale.}
    A SC in SupplyNetPy is a Python graph with node and link
    attributes, so a script can generate and simulate thousands of distinct
    network configurations automatically. This enables building training
    datasets for machine learning models that learn SC behavior as
    a function of network structure and parameters, and is particularly
    attractive for training graph-structured metamodels such as Graph Neural
    Networks (GNNs). Such workflows are difficult or impossible with graphical
    user interface (GUI) based or proprietary-scripting tools.

    \item \textbf{Extensibility.}
    Every entity in SupplyNetPy is a Python class. New node types,
    replenishment policies, supplier selection strategies, and transport
    models are added by subclassing, without modifying the library. The
    current implementation does not yet support continuous material flows
    (e.g., pipelines as links), multi-product shared inventory, or advanced
    logistics features such as fleet management and vehicle routing. These
    can be added incrementally as derived classes. Current scope and planned
    extensions are described in Section~\ref{sec:conclusion}.

    \item \textbf{Flexible integration with the Python ecosystem.}
    Because SupplyNetPy produces standard Python objects, it integrates
    directly with existing Python libraries for optimization (e.g., SciPy),
    visualization (e.g., Matplotlib), machine learning (e.g., PyTorch), and
    GUI development. SimPy's built-in real-time simulation mode supports
    real-time data assimilation and digital twin applications out of the box.
    Users can build application-specific interfaces and workflows directly on
    top of the open, well-documented library.

\end{itemize}


\section{Related Work}
\label{sec:related}

A variety of tools support SC simulation modeling. A recent survey of their use 
across SC application domains can be found in \citep{lone2023open} 
(\href{https://github.com/SupplyChainSimulation/InventOpt/tree/main/review}{GitHub repository}).
The tools compared here are the ones most widely used in SC simulation, organized into three groups that span the
relevant design space: commercial tools (AnyLogic, Arena, Simio, FlexSim, AnyLogistix), general-purpose
open-source simulation libraries (SimPy, Repast), and open-source SC-specific 
libraries (miniSCOT, supplychainpy, SunFlow, Stockpyl).

Commercial simulation tools including AnyLogic \citep{anylogic}, Arena
\citep{arena}, Simio \citep{simio}, and FlexSim \citep{flexsim}
are widely used for modeling and analyzing SCs, particularly for
inventory management, logistics, and resilience. AnyLogic supports DES,
agent-based simulation (ABS), and system dynamics (SD) modeling and is frequently applied to
inventory and logistics problems, leveraging OptQuest for optimization.
AnyLogistix \citep{anylogistix}, developed by the same company, is a
specialized tool for SC simulation and optimization that enables
users to model complex SCs using a visual interface, run network
optimization scenarios, and evaluate system performance under uncertainty.
Arena and FlexSim are commercial DES tools applicable to a broad range of
SC problems. Simio supports both DES and ABS to model SC
aspects such as logistics. These tools provide rich graphical environments,
built-in optimization, and geographic information system (GIS) based visualization, but are proprietary with
platform-specific scripting interfaces.

Open-source libraries such as SimPy \citep{simpy} and Repast
\citep{repast} can model SC aspects including inventory,
logistics, and resilience, and offer greater customization. However, users
must develop all SC components from scratch, requiring significant
programming effort. SimPy is a Python-based DES library, and Repast is primarily
designed for agent-based modeling but also has a Python-based version.

Several open-source libraries target subsets of SC modeling
problems. miniSCOT \citep{miniscot}, developed by Amazon, supports DES
and provides SC-specific components, but lacks comprehensive
documentation and has not been actively maintained since 2021.
\textit{supplychainpy} \citep{supplychainpy} supports Monte Carlo
simulation and automates spreadsheet-based analytical workflows, but does
not support DES. SunFlow \citep{sunflow2020} is a Python-based library
for analytical modeling of SC networks for cost minimization and
network flow optimization, without DES support. Stockpyl \citep{stockpyl}
is an open-source Python library for inventory simulation and analysis using
a time-stepped simulation method, with support for configurable replenishment
policies and node and network level performance metrics. However, it focuses
primarily on analytical inventory optimization and does not support
event-driven simulation or custom replenishment policies via inheritance.
Table~\ref{tab:tools} summarizes the key characteristics of these tools 
alongside SupplyNetPy. The evaluation criteria, represented in the table columns, outline
the key characteristics that define SupplyNetPy's intended scope. To our knowledge, 
there is a need for a well-maintained, open-source Python library combining 
DES with SC-specific components and supporting fully programmatic model 
construction. SupplyNetPy is developed to fill this gap.

\begin{table}[htb]
\centering
\caption{Comparison of simulation modeling tools.\label{tab:tools}}
\footnotesize
\setlength{\tabcolsep}{4pt}
\begin{tabular}{|>{\raggedright\arraybackslash}p{0.13\linewidth}|>{\centering\arraybackslash}p{0.07\linewidth}|>{\centering\arraybackslash}p{0.1\linewidth}|>{\centering\arraybackslash}p{0.05\linewidth}|>{\centering\arraybackslash}p{0.05\linewidth}|>{\centering\arraybackslash}p{0.05\linewidth}|>{\centering\arraybackslash}p{0.07\linewidth}|>{\centering\arraybackslash}p{0.1\linewidth}|}
\hline
\textbf{Tool} & \textbf{Open Source} & \textbf{SC-Specific Components} & \multicolumn{3}{c|}{\textbf{Simulation Methods}} & \textbf{GIS Support} & \textbf{Performance Reporting} \\
\hline
 & & & \textbf{DES} & \textbf{ABS} & \textbf{SD} & & \\
\hline
\textbf{SupplyNetPy} & \greencheck & \greencheck & \greencheck & No & No & No & \greencheck \\
\hline
miniSCOT      & \greencheck & \greencheck & \greencheck & No & No & No & No \\
\hline
supplychainpy & \greencheck & \greencheck & No & No & No & No & \greencheck \\
\hline
SunFlow       & \greencheck & \greencheck & No & No & No & No & \greencheck \\
\hline
Stockpyl      & \greencheck & \greencheck & \multicolumn{3}{c|}{Time-stepped} & No & \greencheck \\
\hline
SimPy         & \greencheck & No & \greencheck & No & No & No & No \\
\hline
Repast        & \greencheck & No & \greencheck & \greencheck & \greencheck & No & No \\
\hline
Arena         & No & No & \greencheck & \greencheck & \greencheck & \greencheck & \greencheck \\
\hline
AnyLogic      & No & No & \greencheck & \greencheck & \greencheck & \greencheck & \greencheck \\
\hline
AnyLogistix   & No & \greencheck & \greencheck & \greencheck & \greencheck & \greencheck & \greencheck \\
\hline
Simio         & No & No & \greencheck & \greencheck & No & \greencheck & \greencheck \\
\hline
FlexSim         & No & No & \greencheck & No & No & \greencheck & \greencheck \\
\hline
\end{tabular}
\end{table}


\section{SupplyNetPy: Architecture and Implementation}
\label{sec:library}

SupplyNetPy is a Python library consisting of well-designed and pre-validated component classes
that can be configured and connected together to describe complex SC networks with arbitrary structures.
This raises the modeling abstraction from building and debugging a simulation model to simply describing the structure and attributes of a chain.
The library's component set, modeled parameters, inventory replenishment policies, and performance metrics were guided by a structured review 
of recent SC simulation literature. An initial version appeared in our earlier InventOpt work \citep{lone2023open}, and 
a more comprehensive discussion will appear in a forthcoming extended version of this work.
\begin{figure}[htb]
\centering
\includegraphics[width=0.48\linewidth]{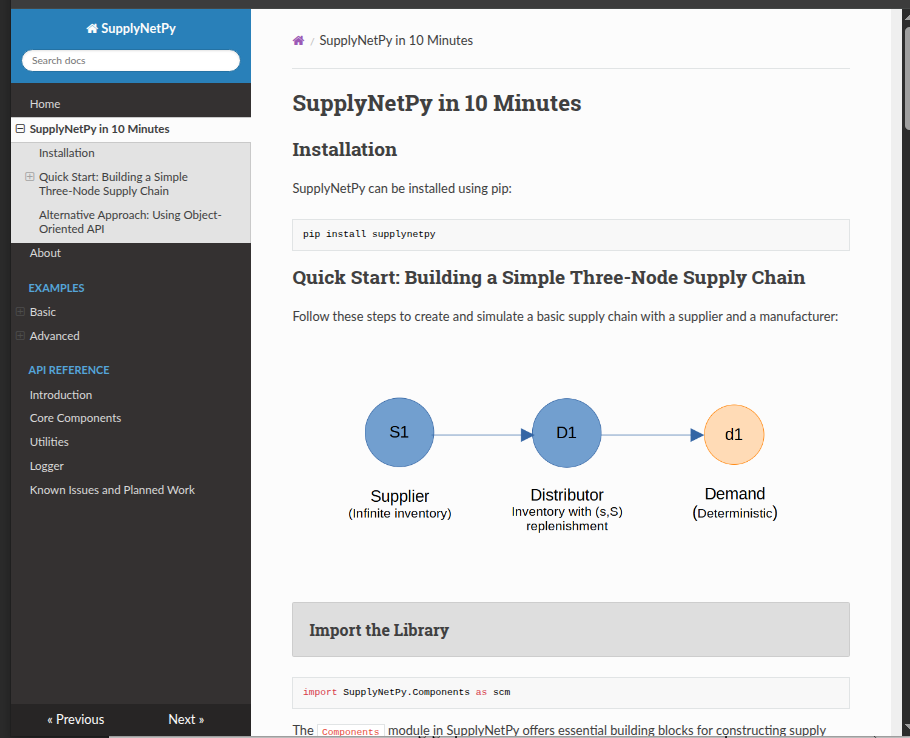}
\hfill
\includegraphics[width=0.48\linewidth]{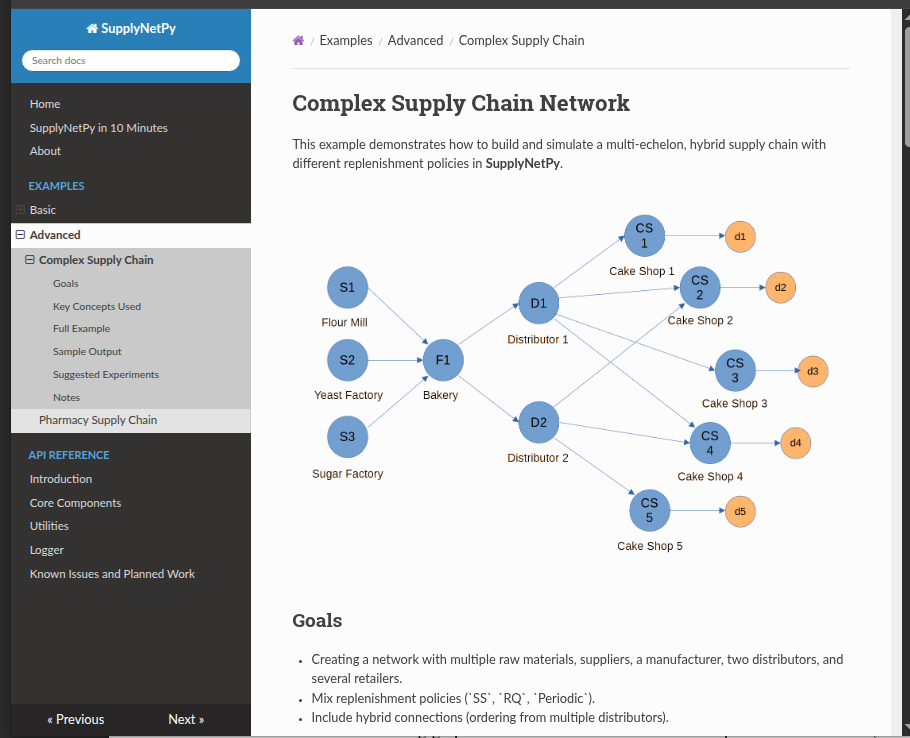}
\caption{Screenshots of the SupplyNetPy online documentation
showing the API guide and usage examples\\ URL: \url{https://supplychainsimulation.github.io/SupplyNetPy}.\label{fig:snpdoc}}
\end{figure}

\subsection{Architecture and Components}
\label{subsec:arch}

A SC in SupplyNetPy is modeled as a directed graph with two
fundamental objects: \textit{nodes} and \textit{links}. Nodes represent
SC entities and fall into three broad categories: source nodes
(manufacturers and suppliers that produce or supply goods), intermediate
inventory-holding nodes (distributors and retailers that stock and forward
product), and leaf demand nodes that consume product and have no downstream
connections. Well-defined structural rules govern the graph: a supplier
has no upstream inventory source, and a demand node has no downstream sink.
The graph need not be a tree; an inventory node can be connected to multiple
upstream manufacturers or suppliers, with a configurable supplier selection
policy to choose among them. Links model directed transport edges between
nodes.
A detailed application programming interface (API) guide describing all classes and their hierarchy is available
in the official documentation. The main classes are as follows:

\begin{itemize}

    \item \texttt{Node}: the base class for all SC entities.
    Stores common attributes: unique identifier, name, node type, and
    geographic location. Supports stochastic disruption modeling via a
    configurable failure probability (\texttt{failure\_p}) and Python
    callables for disruption duration and recovery time.

    \item \texttt{Supplier}: models a source of raw materials or finished
    goods. Supports finite and infinite inventory modes. Default behavior
    mines or produces items at a configurable rate.

    \item \texttt{Manufacturer}: extends \texttt{Node} with manufacturing
    behavior. Orders raw material from a connected \texttt{Supplier},
    assembles products in configurable batch sizes, and holds finished-goods
    inventory for downstream distribution.

    \item \texttt{InventoryNode}: models any intermediate node that holds
    inventory and replenishes from upstream suppliers. Used to instantiate
    distributors and retailers. Manages inventory, serves downstream orders,
    and places replenishment orders according to a configured policy.

    \item \texttt{Demand}: models external customer demand. Accepts Python
    callables for inter-arrival time and order quantity, enabling demand
    from any probability distribution.

    \item \texttt{Link}: models a directed transport edge between two nodes.
    Attributes include source node, sink node, transportation cost, and lead
    time (specified as a Python callable for deterministic or stochastic lead
    times). Also supports a link-level failure probability for disruption
    modeling.

    \item \texttt{Inventory}: manages inventory held at a node. Supports
    both non-perishable and perishable inventory. For perishable inventory,
    each unit is tagged with its manufacture date and shelf life, and the expired
    items are removed using a FIFO discipline. Tracks holding costs, shortage
    costs, and waste costs due to expiry.

    \item \texttt{Product}: represents a finished good with attributes
    including manufacturing cost, sell price, production time, and raw
    material requirements.

    \item \texttt{RawMaterial}: represents an input material with extraction
    quantity, extraction time, and cost attributes.

\end{itemize}

\noindent \textbf{Dual API.}
SupplyNetPy offers two interfaces for building and running models.
The \textit{functional API} uses the helper functions
\texttt{create\_sc\_net()} and \texttt{simulate\_sc\_net()}, which accept
dictionaries describing nodes, links, and demand processes and handle all
SimPy environment setup internally. The \textit{object-oriented API} exposes
the underlying SimPy environment directly, allowing users to instantiate
node and link objects manually, attach additional SimPy processes, and
control the simulation loop. Both APIs produce the same underlying simulation
model and performance reports. A minimal example using the functional API
is shown in Listing~\ref{lst:minimal}.

\subsection{Replenishment Policies and Supplier Selection}
\label{subsec:policies}

Three families of inventory replenishment policies are provided, each
implemented as a class derived from the abstract
\texttt{InventoryReplenishment} base:

\begin{itemize}
    \item \textbf{(s, S) min-max policy} (\texttt{SSReplenishment}):
    when inventory position falls to or below reorder level $s$, an order
    is placed to bring the position up to $S$.

    \item \textbf{(R, Q) reorder-quantity policy} (\texttt{RQReplenishment}):
    a fixed quantity $Q$ is ordered whenever inventory position falls to or
    below reorder point $R$.

    \item \textbf{Periodic review policy} (\texttt{PeriodicReplenishment}):
    the inventory is reviewed every $T$ time units and a fixed quantity $Q$
    is ordered at each review.
\end{itemize}

\noindent All three policies support an optional safety stock parameter.
New policies are added by subclassing \texttt{InventoryReplenishment} and
overriding the \texttt{run()} method.

When a node has multiple upstream suppliers, a supplier selection strategy
determines which supplier fulfills each replenishment order. Four strategies
are provided: \texttt{SelectFirst} (always use the first listed supplier),
\texttt{SelectAvailable} (first supplier with sufficient stock),
\texttt{SelectCheapest} (minimize transportation cost), and
\texttt{SelectFastest} (minimize lead time). Custom strategies are added
by subclassing \texttt{SupplierSelectionPolicy}.

\subsection{Key Features}
\label{subsec:features}

\begin{itemize}

    \item \textbf{Perishable and non-perishable inventory.}
    Each inventory item carries a manufacture date, and items are removed from
    stock when their shelf life is exceeded, using a FIFO discipline. Waste
    costs due to expiry are tracked alongside holding and shortage costs.
    Applicable to cold chains, pharmaceutical SCs, and food
    distribution networks.

    \item \textbf{Node and link disruption modeling.}
    Nodes are configured with failure probability and callable disruption
    and recovery durations. Disruptions propagate naturally via unfulfilled
    orders and inventory shortfalls, enabling resilience studies and what-if
    analyses.

    \item \textbf{Stochastic demand and lead times.}
    Demand arrival times, order quantities, and link lead times are all
    specified as Python callables, enabling any distribution (Poisson,
    normal, empirical, or user-defined).

    \item \textbf{Performance reporting and logging.}
    Each node exposes a \texttt{statistics} object with node-level metrics:
    inventory levels over time, demand placed and fulfilled, shortages,
    waste, holding cost, transportation cost, revenue, and profit.
    Network-level aggregates are computed by \texttt{simulate\_sc\_net()}.
    Detailed simulation event logs are also generated.

    \item \textbf{Extensibility.}
    Any class in the hierarchy can be subclassed to implement
    problem-specific behavior without altering the library code.
    Although the current version does not support 
    multi-product shared inventory or advanced logistics
    features such as fleet management and vehicle routing, these are planned
    extensions and can be added by subclassing the node and link classes. 

\end{itemize}

\subsection{A Minimal Example}
\label{subsec:example}

Listing~\ref{lst:minimal} shows a complete, runnable three-node SC
model using the functional API: an infinite supplier, a distributor
with a min-max replenishment policy, and a Poisson demand process. The
model is assembled in under fifteen lines with no SimPy boilerplate.

\begin{lstlisting}[language=Python,
    caption={A minimal three-node supply chain in SupplyNetPy.},
    label={lst:minimal}]
import SupplyNetPy.Components as scm
import random

nodes = [
  {'ID':'S1', 'name':'Supplier',
   'node_type':'infinite_supplier'},
  {'ID':'D1', 'name':'Distributor',
   'node_type':'distributor',
   'capacity':200, 'initial_level':100,
   'inventory_holding_cost':0.5,
   'replenishment_policy': scm.SSReplenishment,
   'policy_param': {'s':60, 'S':200},
   'product_buy_price':10,
   'product_sell_price':15}
]
links   = [{'ID':'L1','source':'S1','sink':'D1',
            'cost':5,
            'lead_time': lambda: random.expovariate(1/3)}]
demands = [{'ID':'d1','name':'Demand',
            'order_arrival_model':
              lambda: random.expovariate(1),
            'order_quantity_model': lambda: 10,
            'demand_node':'D1'}]

net = scm.create_sc_net(nodes, links, demands)
net = scm.simulate_sc_net(net, sim_time=360)
stats = net['nodes']['D1'].stats.get_statistics()
print(stats)
\end{lstlisting}

\noindent To model perishable inventory, add
\texttt{'inventory\_type':'perishable'} and \texttt{'shelf\_life':5} to
the distributor dictionary. To model node disruptions, add
\texttt{'failure\_p':0.01} and callable disruption and recovery time
functions. The online documentation provides several basic and advanced usage examples.

\subsection{Implementation and Release History}
\label{subsec:history}

SupplyNetPy has been under continuous development since 2022. An initial
prototype with a limited feature set, then named InventOpt, was first
published in \citep{lone2023open}. The library was subsequently expanded,
improved, and thoroughly validated \citep{lone2024development}, and was released publicly on GitHub under
an MIT license in 2025 as SupplyNetPy. It is currently in active use in
research projects for programmatic generation of SC models as
training datasets for machine learning methods.


\section{Validation}
\label{sec:validation}

Validation of SupplyNetPy was carried out through the following complementary approaches:

\begin{enumerate}

    \item \textbf{Unit tests.} Individual components were tested using Python \textit{pytest} to verify correct isolated behavior.

    \item \textbf{Comparison against analytical benchmarks.} For well-known inventory problems with closed-form solutions, simulated outputs were compared against the analytical optimum, confirming that component behavior is theoretically sound.

    \item \textbf{Component-wise comparison against a commercial tool.}
    Small, deterministic SC models were constructed identically in SupplyNetPy and AnyLogistix, and numerical outputs were compared component by component. The goal was to validate correct implementation of each component and replenishment policy, not to compare features or performance. A good match was found for most metrics, and the observed differences and their underlying causes are well understood and are discussed in Section~\ref{subsec:anylogistix}.

    \item \textbf{Comparison against published case study results.}
    SupplyNetPy was used to reproduce a pharmacy SC case study \citep{czerniak2021improving} for which the full problem specification, parameters, and results were available, validating the library on a realistic scenario that exercises advanced features (especially perishability) simultaneously. This is presented in Section~\ref{sec:casestudy}.

\end{enumerate}

\subsection{Comparison Against Analytical Results}
\label{subsec:analytical}

Analytical methods provide closed-form or benchmark results for simple,
canonical inventory configurations. We validated against three classic
textbook examples \citep{chopra2007supply}. For each, the full detailed description,
model code and plots are available in the SupplyNetPy online documentation.
All stochastic results below are reported as the mean over independent
simulation replications together with a 95\% confidence interval (CI), computed using 
Python libraries (NumPy and SciPy). Each simulated estimate is compared to its analytical benchmark, 
and any discrepancies are examined and explained.
\begin{enumerate}

\item \textbf{Newsvendor problem} \\
(\url{https://supplychainsimulation.github.io/SupplyNetPy/example-newvendor})\\
A vendor orders $Q$ units at cost $c$ per unit
and sells them at price $p$ (unsold units salvaged at value $s$) with normally
distributed demand ($\mu$, $\sigma$), and the profit-maximizing order quantity
$Q^*$ is given by a closed-form expression. We implemented this as a three-node
SC: an infinite-capacity supplier, a newsvendor node with
perishable inventory (shelf life of one period, periodic replenishment), and
a demand node. Setting $c=2$, $p=5$, $s=1$, $\mu=100$, $\sigma=15$ gives
$Q^* \approx 110$ units analytically. We swept $Q$ from 10 to 200, ran 1{,}000
simulation replications at each value, and verified that the simulated profit
curve peaks at $Q \approx 110$, matching the analytical optimum $Q^* \approx 110$.
At this optimum the mean simulated profit is $278.2$ with a 95\% CI of $[273.5, 282.8]$. This confirms correct
perishable inventory handling and waste cost accounting.

\item \textbf{Economic Order Quantity (EOQ)} \\
(\url{https://supplychainsimulation.github.io/SupplyNetPy/EOQ_est})\\
An inventory system with annual demand of 12{,}000 units, fixed order cost \$4{,}000 per lot, unit
cost \$500, and annual holding cost of 20\%. The analytical EOQ is approximately 980 units. Over a long horizon 
(4{,}000 days), the simulated cost-minimizing lot size was approximately 1{,}010 units, within 3\% of 
the analytical value. Because the EOQ total-cost curve is very flat near its optimum (the cost penalty 
at 1{,}010 versus 980 units is under 0.1\%), this agreement validates the (R, Q) replenishment policy 
implementation and cost tracking.

\item \textbf{Safety stock estimation} \\
(\url{https://supplychainsimulation.github.io/SupplyNetPy/safety_inv_est})\\
We modeled an inventory node with normally distributed weekly demand ($\mu = 2{,}500$, $\sigma = 500$) and a two-week
replenishment lead time, with reorder point 6{,}000 and order quantity
10{,}000. Over 100 simulation replications, the estimated safety stock level
was 1{,}346.8 units (95\% CI $[1{,}335.9, 1{,}357.6]$), the average
inventory level was 6{,}311.7 units (95\% CI $[6{,}303.1, 6{,}320.4]$),
and the average order flow time was 13.81 days (95\% CI $[13.79, 13.83]$).
The simulated safety stock and average inventory lie above their analytical
values of 1{,}000 and 6{,}000 units, whereas the flow time lies below its
analytical value of 2.4 weeks (16.8 days). These expected deviations arise
because resampling negative demand realizations (an artifact of the normal
approximation) makes the effective demand higher than the nominal normal
demand assumed analytically, pushing the two inventory measures above and
the flow time below their respective analytical values.
\end{enumerate}

\subsection{Component-Wise Validation Against a Commercial Tool}
\label{subsec:anylogistix}

The purpose of this validation was \emph{not} to compare features or
performance against a commercial tool, but to check that each SupplyNetPy
component and replenishment policy was implemented correctly by verifying
that numerical results match a trusted reference for a structurally
simple, deterministic model. We chose AnyLogistix (free, personal learning edition) 
as the reference as it provides several SC-specific components with
well-documented behavior. We selected a structurally minimal configuration:
a single inventory node (a distribution center) supplied by an infinite-capacity supplier, with deterministic demand. 
This isolated each component and policy from any network-induced effects. 
SupplyNetPy currently supports a subset of replenishment policies that are available in AnyLogistix.
All components and policies currently available in SupplyNetPy were carefully covered in 
the step-by-step matching and validation process: nodes (supplier, distributor, retailer, and demand),
links (transportation costs and lead times), and inventory (all replenishment
policies and cost accounting). Simulations were run over a long duration to
accumulate sufficient statistics. The full model configurations, tested parameter ranges, 
and numerical results of this validation study are available here: \url{https://github.com/SupplyChainSimulation/SupplyNetPy/tree/main/validation}.

\textbf{Results Summary:} 
For most deterministic model configurations (non-stochastic demand and constant lead times), 
SupplyNetPy and AnyLogistix produced identical results across all tracked
metrics for both single-echelon and two-echelon configurations over long simulations.
This confirmed that the core inventory management logic, replenishment triggering, and cost
accounting were correctly implemented across all components and policies. 

In the few non-matching instances, the exercise surfaced cases where we
had assumed a subtly different interpretation of a model parameter or policy.
These were subsequently re-implemented in a more flexible, configurable manner,
allowing the modeler to precisely specify the desired variant. Output metrics
matched fully under one of the chosen variants, with the alternatives offering
flexibility beyond the default assumptions of the commercial tool. 
The following are some specific examples of such parameters/policies where subtle variations are possible
and have been added in SupplyNetPy:
\begin{enumerate}
    \item ~In a three-echelon SC with supplier, factory, and distributor nodes, AnyLogistix defaults to zero transportation cost on the factory-distributor link. SupplyNetPy generalizes this assumption into a configurable parameter, allowing users to specify either zero or nonzero values.
    \item ~A parameter \texttt{consume\_available} has been introduced to interpret demand generated at a single node as individual demand. When the required order quantity is not fully available, the model consumes whatever inventory exists. In AnyLogistix, achieving this behavior requires creating separate demand nodes.
\end{enumerate}

We found a few \textbf{pathological configurations} where results differed
even though both models were deterministic and the implementation was correct.
The causes are as follows:

\textbf{(A) Simultaneous event ordering in the simulation core:} when demand
inter-arrival times and lead times are integer multiples of each other,
combined with a no-backorder policy (unsatisfied demand is lost immediately),
a replenishment arrival and a demand event can coincide exactly at time $t$,
and the outcome (demand fulfilled or lost) depends on which event is processed
first. Each tool resolves such ties deterministically but differently:
SupplyNetPy uses an event ID assigned at event creation time, whereas
AnyLogistix uses the order in which processes register events. Event logs
diverge at the first such collision, and the discrepancy accumulates over
long runs. The issue does not arise under a backorder policy (queued demand
is fulfilled at the same simulation time once replenishment is processed),
and disappears in stochastic configurations, where results of the two tools
match in a statistical sense.

\textbf{(B) Floating-point precision divergence in real-valued deterministic
configurations:} when time intervals are non-integer real values (for example,
a link lead time of $1.3\overline{3}$), floating-point representation errors
accumulate differently across the two implementations (depending on the
underlying language), slightly shifting effective event timestamps over long
runs and altering the ordering of near-simultaneous events, with the same
consequences and the same stochastic-case resolution as in (A).

There is no single universally correct scheme for simultaneous-event handling
in DES, and both tools behave correctly according to their own documented
semantics. Real-world SC models are stochastic, with results
estimated over multiple replications, which averages out such
boundary-condition differences. Discrepancies were thus confined to specific,
pathological boundary cases, and the component-wise comparison confirmed
correct (with respect to the commercial tool) implementation of all
components and policies.

\subsection{Arbitrary Network Graphs (Bakery Supply Chain Example)}
\label{subsec:bakery}

A bakery SC example in the documentation (see Figure~\ref{fig:snpdoc}, right)
illustrates the modeling of arbitrary multi-echelon SC graphs. Unlike the
minimal three-node example in Section~\ref{subsec:example}, this network involves
multiple echelons, several interconnected node types, and perishable inventory. It
was used to verify that complex topologies, multi-echelon event scheduling, and
network-level performance aggregation are correctly handled. Performance on large
networks is assessed separately in Section~\ref{sec:casestudy}.


\section{Case Study and Performance Evaluation}
\label{sec:casestudy}

\subsection{Case Study: Pharmacy Supply Chain}
\label{subsec:pharmacy}

After component-wise and policy-wise unit validation and comparison with an
existing tool as reference, our goal was to validate the unique features
supported by SupplyNetPy (such as perishability) through a comprehensive
case study for which published results were available. We selected a
pharmacy SC case study published at Winter Simulation Conference
2021 \citep{czerniak2021improving}, as it details the full problem
setup and results, allowing us to implement the exact same model in
SupplyNetPy and directly compare the outputs. The full implementation,
code, and results are available in the SupplyNetPy documentation
(\url{https://supplychainsimulation.github.io/SupplyNetPy/case_study_pharma/}).

\noindent\textbf{Problem description:}
The system is a single-echelon pharmacy SC: a pharmaceutical
supplier with stochastic disruptions supplies a hospital pharmacy that
faces stochastic Poisson demand. The pharmacy stocks a perishable drug
with a finite shelf life. Expired units are discarded and unmet demand is
lost. Replenishment follows a periodic $(s, S)$ policy with a fixed lead
time. The objective is to find the $(s, S)$ combination that minimizes
expected total daily cost, which comprises shortage, waste, holding, and
ordering components. All parameter values are provided in the online
documentation.

\begin{figure}[htb]
\centering
\includegraphics[width=0.48\linewidth]{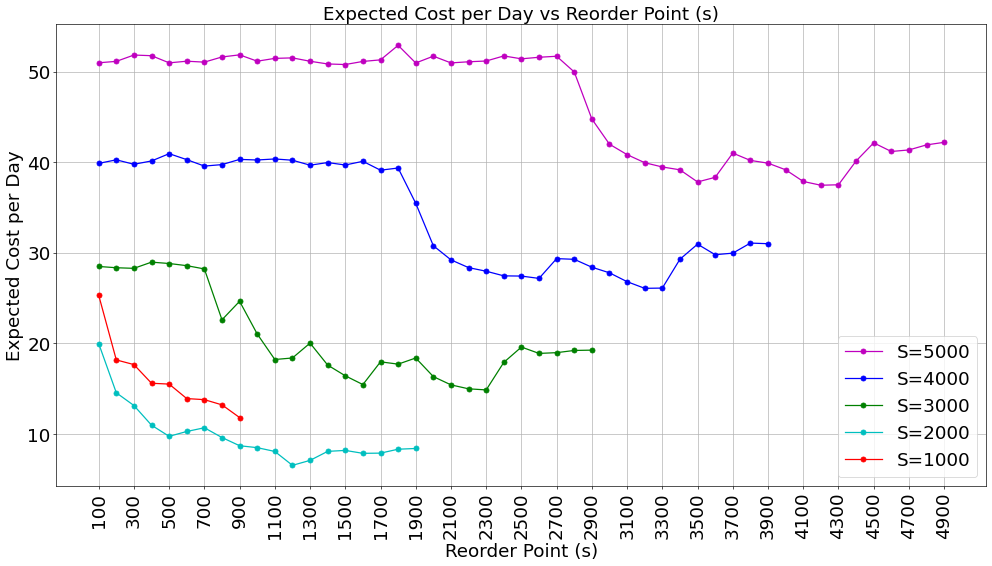}
\hfill
\includegraphics[width=0.48\linewidth]{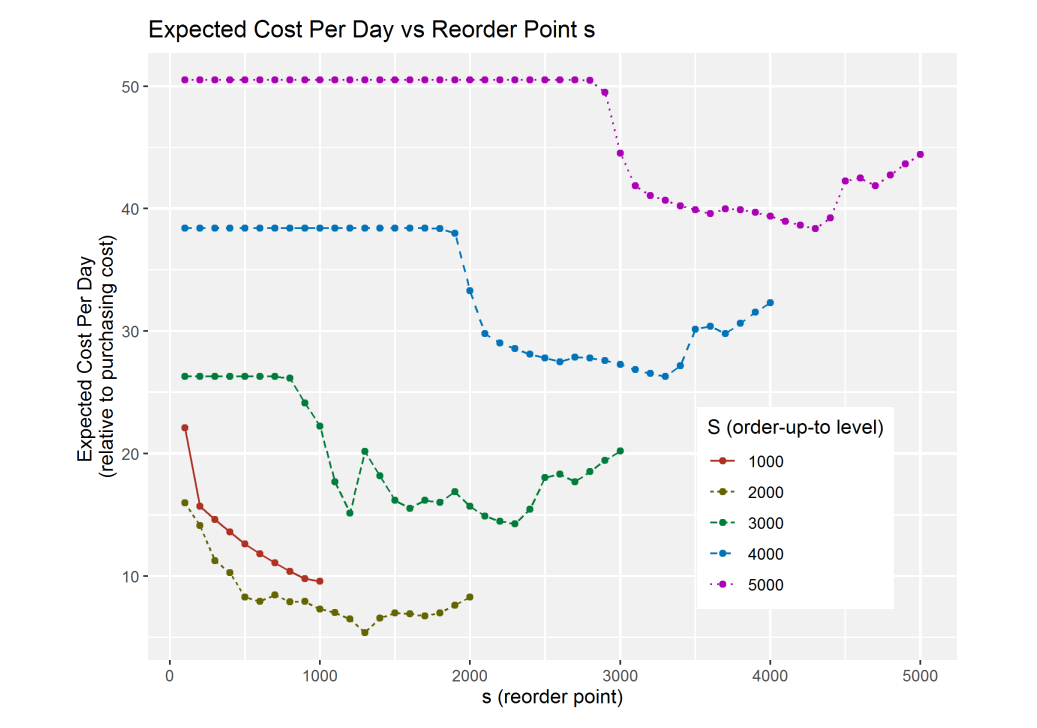}
\caption{Expected daily cost vs.\ $(s, S)$ parameters. Left: results
obtained with SupplyNetPy. Right: original results from
\protect\citep{czerniak2021improving}. The optimal policy region is
consistent across both.\label{fig:pharma_results}}
\end{figure}

\noindent\textbf{Implementation in SupplyNetPy:}
The model was implemented as a three-node SC. The supplier node
was configured with stochastic disruption using Python callables for
disruption onset and recovery duration. The pharmacy was modeled as an
\texttt{InventoryNode} with \texttt{inventory\_type=`perishable'}, with
the $(s, S)$ policy set directly via node parameters. Batch-level
expiration was handled by passing a \texttt{manufacturer\_date\_cal()}
callable that assigns a manufacture date to each incoming shipment,
enabling FIFO expiry tracking. The demand node generates Poisson-distributed
daily arrivals. An exhaustive grid search over $(s, S)$ was performed,
running 1{,}000 replications per parameter combination (chosen based on a
convergence analysis of the standard error on mean daily cost). We present a few of the 
interesting results obtained below:

\noindent\textbf{Results:}
As an example illustrating the match between the original study and the 
model replicated using SupplyNetPy, Figure~\ref{fig:pharma_results} shows the expected daily cost surface
obtained from SupplyNetPy alongside the original result from \citep{czerniak2021improving}. 
The plots show that for low order-up-to levels the cost is driven by frequent shortages, 
while for high levels waste cost dominates, producing
a well-defined optimal band in the $(s, S)$ space. Further, the optimal region
identified by SupplyNetPy is consistent with the original study. 
Similar experiment runs across other configurations match the observations reported in 
the original pharmacy study. The independent
replication of this case study's results confirms that SupplyNetPy 
correctly models perishable inventory with FIFO expiry,
stochastic demand, probabilistic supply disruptions, and the
associated cost structure.

\subsection{Performance Evaluation}
\label{subsec:performance}

We evaluated how SupplyNetPy's single-run execution time scales with
SC size $N$ and simulation length. Parameterized networks were
generated programmatically: for a given $N$, one-fifth of nodes are
suppliers, one-tenth manufacturers, one-seventh distributors, and the
remainder retailers, approximating a realistic multi-echelon structure.
All suppliers connect to all manufacturers, manufacturers to all
distributors, and distributors to all retailers (no upstream links,
ensuring an acyclic flow). All suppliers have infinite capacity. Stochastic
demand with exponential inter-arrival times is generated at each retailer.
For each $N$, 30 independent replications of length 500 time units were
executed to record the distribution of execution time. The network topology and all node
and link attributes remained fixed, with only the simulation random-number stream
varied across replications, eliminating the additional variability that a randomly
regenerated network would otherwise introduce.
A separate experiment fixed $N=50$ and varied simulation length from 500 to 
5{,}000 time units. Replications are independent and were executed in parallel 
on a dual-socket Intel Xeon Gold 6148 server (two 20-core CPUs at 2.40\,GHz, 
80 logical cores) running 64-bit Linux.

\begin{figure}[htb]
\centering
\includegraphics[width=0.48\linewidth]{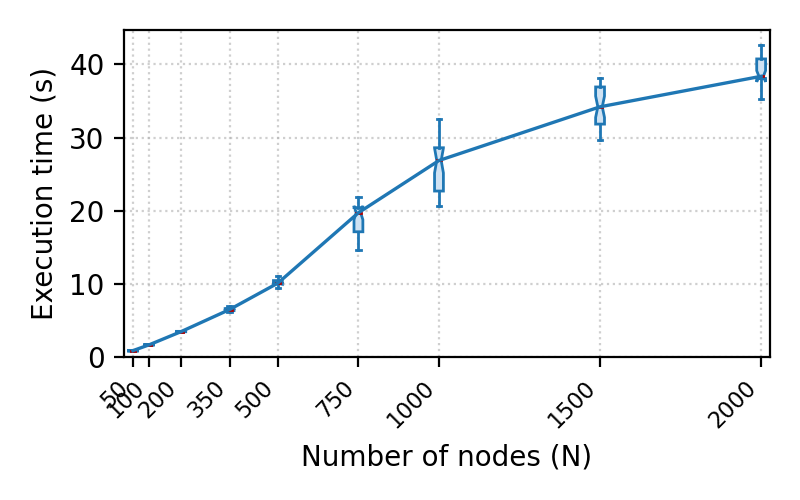}
\hfill
\includegraphics[width=0.48\linewidth]{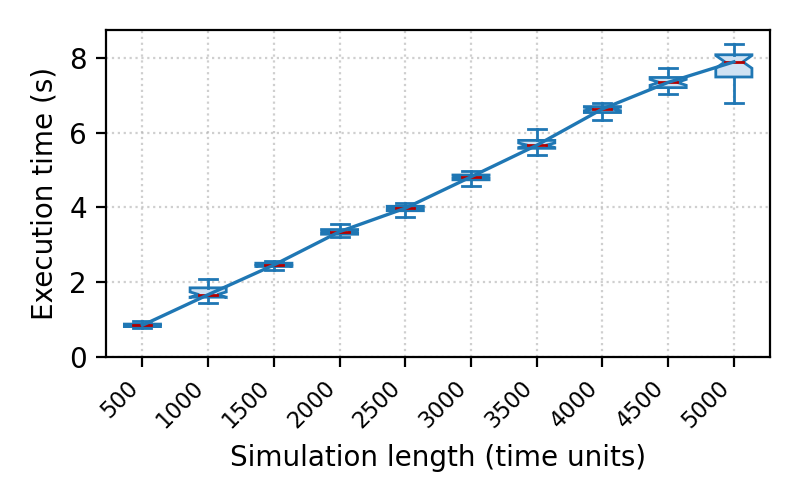}
\caption{Execution time scaling.
Left: vs.\ number of nodes $N$ (fixed simulation length 500). Right: vs.\
simulation length (fixed $N=50$). Each box spans the interquartile range,
the notch marks the 95\% confidence interval of the median, and the line
connects the per-configuration medians.\label{fig:performance}}
\end{figure}

Figure~\ref{fig:performance} reports the distribution of execution time
across replications. Execution time grows linearly with simulation length, with
tight median confidence intervals throughout. Scaling with $N$ is close to
linear, and the per-event cost grows only gradually as $N$
increases, so SupplyNetPy introduces no super-linear overhead from component
management or statistics collection. The wider median confidence intervals seen
for mid-range $N$ do not reflect structural variability. They arise from ordinary 
run-to-run timing jitter (process scheduling and memory effects under parallel load), 
while the medians themselves remain stable. Overall, the results confirm practical
usability for replication-based studies on realistically sized networks.
The performance study and these plots serve to provide a rough idea of the
simulation speed offered by SupplyNetPy. Further improvements may be possible
by running the simulation under a compiled mode (using PyPy), and by
supporting parallel simulation of SupplyNetPy models by model partitions and
using parallel and distributed discrete-event simulation (PDES) paradigms,
which are future development directions. Enabling parallel simulation could
make it possible to exploit multi-core or cluster systems for fast design
space exploration and real-time digital twins of massive global-scale SC
models.

\section{Conclusions and Future Work}
\label{sec:conclusion}

Supply chain simulation is critical for the analysis, optimization, and
resilience assessment of modern SCs. Despite this importance,
the open-source software ecosystem lacks well-maintained, DES-based
libraries that provide SC-specific components and support fully
programmatic model construction. This paper presented SupplyNetPy, an
open-source Python library that addresses this gap.

SupplyNetPy provides a modular, component-based architecture that raises 
the modeling abstraction for SCs and facilitates easy lego-blocks style 
modeling, validation, and extensive performance reporting. Key features 
include: (i) support for modeling perishable inventory, 
(ii) node disruption modeling for resilience assessment, (iii) several
types of built-in replenishment policies and supplier selection strategies,
all extensible via class inheritance, and (iv) a dual API enabling both rapid
functional-style prototyping and deep customization.

While the library currently supports both discrete and real-valued inventory, 
the inventory transport is modeled as discrete events. Support for continuous
flows (pipelines) is planned in future versions. Further, planned extensions include:
(1) a more comprehensive node and link disruptions modeling and built-in resilience metrics reporting,
(2) more link types such as fleets, supporting carbon cost reporting,
(3) support for multi-product shared inventory, and bundled orders and shipments,
(4) scalable real-time simulation support for digital twin applications by applying parallel execution strategies, and
(5) metamodel support for optimization, particularly exploring graph-appropriate metamodels such as GNNs.

This paper presented the motivation, design, and architecture of SupplyNetPy, 
along with detailed validation results (comparing with analytical results, a reference commercial tool, and a published case study) as well as performance results. SupplyNetPy is developed as an academic project and is freely available as an open-source library under an MIT license.

\section*{Acknowledgments}
The authors thank Lekshmi~P for her contributions to the library documentation
and initial validation tests, and Divyansh~Pandey for his work during a student
internship on validation and on the development of SupplyNet Web, a web-based
GUI for SupplyNetPy, available at \url{https://supply-net-web.vercel.app/}.

\bibliographystyle{plain}
\bibliography{bibliography}

\end{document}